\documentclass[letterpaper, 10 pt, conference]{ieeeconf}  % Comment this line out if you need a4paper

\IEEEoverridecommandlockouts                              % This command is only needed if 
\title{\LARGE \bf Markerless Motion Capture and Biomechanical Analysis Pipeline}

\usepackage{amsmath}
\usepackage{amsfonts}
\usepackage{booktabs}
\usepackage{xcolor}
\usepackage{multirow}
\usepackage[figuresright]{rotating}
\RequirePackage[colorlinks=true, allcolors=blue]{hyperref}

\usepackage[
    backend=biber,
    style=ieee,
    natbib=true,
    doi=false,
    url=false,
    isbn=false,
]{biblatex}

\AtEveryBibitem{%
   \clearfield{Title}%
   \clearfield{eprint}%
   \clearfield{note}%
}

\newbox{\myorcidaffilbox}
\sbox{\myorcidaffilbox}{\large\includegraphics[height=1.1ex]{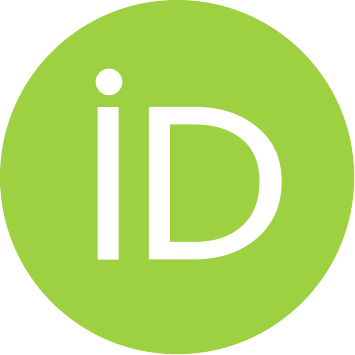}}
\newcommand{\orcidaffil}[1]{%
  \href{https://orcid.org/#1}{\usebox{\myorcidaffilbox}}}

\author{R. James Cotton$^{1,2,*,\orcidaffil{0000-0001-5714-1400}}$, Allison DeLillo$^1$, Anthony Cimorelli$^1$, Kunal Shah$^1$, \\ J.D. Peiffer$^{1,3,\orcidaffil{0000-0003-2382-8065}}$, Shawana Anarwala$^1$, Kayan Abdou$^1$, Tasos Karakostas$^{1,2}$ 
\thanks{This work was generously supported by the Research Accelerator Program of the Shirley Ryan AbilityLab and with funding from the Restore Center P2C (NIH P2CHD101913). We also thank István Sárándi for advice obtaining confidence estimates from MeTrABs.}
\thanks{*rcotton@sralab.org, 1. Shirley Ryan AbilityLab, 2. Department of Physical Medicine and Rehabilitation,  Northwestern University, 3. Department of Biomedical Engineering, Northwestern University}
}

\begin{document}

\maketitle

\begin{abstract}
Markerless motion capture using computer vision and human pose estimation (HPE) has the potential to expand access to precise movement analysis. This could greatly benefit rehabilitation by enabling more accurate tracking of outcomes and providing more sensitive tools for research. There are numerous steps between obtaining videos to extracting accurate biomechanical results and limited research to guide many critical design decisions in these pipelines. In this work, we analyze several of these steps including the algorithm used to detect keypoints and the keypoint set, the approach to reconstructing trajectories for biomechanical inverse kinematics and optimizing the IK process. Several features we find important are: 1) using a recent algorithm trained on many datasets that produces a dense set of biomechanically-motivated keypoints, 2) using an implicit representation to reconstruct smooth, anatomically constrained marker trajectories for IK, 3) iteratively optimizing the biomechanical model to match the dense markers, 4) appropriate regularization of the IK process. Our pipeline makes it easy to obtain accurate biomechanical estimates of movement in a rehabilitation hospital.
\end{abstract}

\section*{Introduction}\label{introduction}

Easy-to-use tools for routine and accurate movement analysis, such as gait analysis, are critical for rehabilitation. They can enable more precise monitoring of clinical outcomes and power more sensitive research studies. The clinical gold standard is marker-based optical motion capture \cite{winter_biomechanics_2009}, but the time and cost associated with this limit its utilization. Recently, advances in computer vision and human pose estimation (HPE) \cite{zheng_deep_2020} have enabled markerless motion capture approaches from multiple calibrated cameras
\cite{nakano_evaluation_2020, uhlrich_opencap_2022, kanko_concurrent_2020, kanko_inter-session_2021, matthis_jonathan_samir_2022_7233714}, potentially expanding access to precise movement analysis. 

Markerless motion capture typically involves two key steps: (1) reconstructing 3D keypoint (i.e., virtual marker) trajectories from the HPE output of multiple calibrated and synchronized cameras and (2) performing inverse kinematic (IK) fits of a biomechanical model to these trajectories. IK includes scaling the individual's biomechanical model and computing the joint kinematics that best aligns with the trajectories. An additional step may compute the kinetics, or joint torques and ground reaction forces, but we do not consider that in this work.

\begin{figure}
    \centering
    \includegraphics[width=0.95\linewidth]{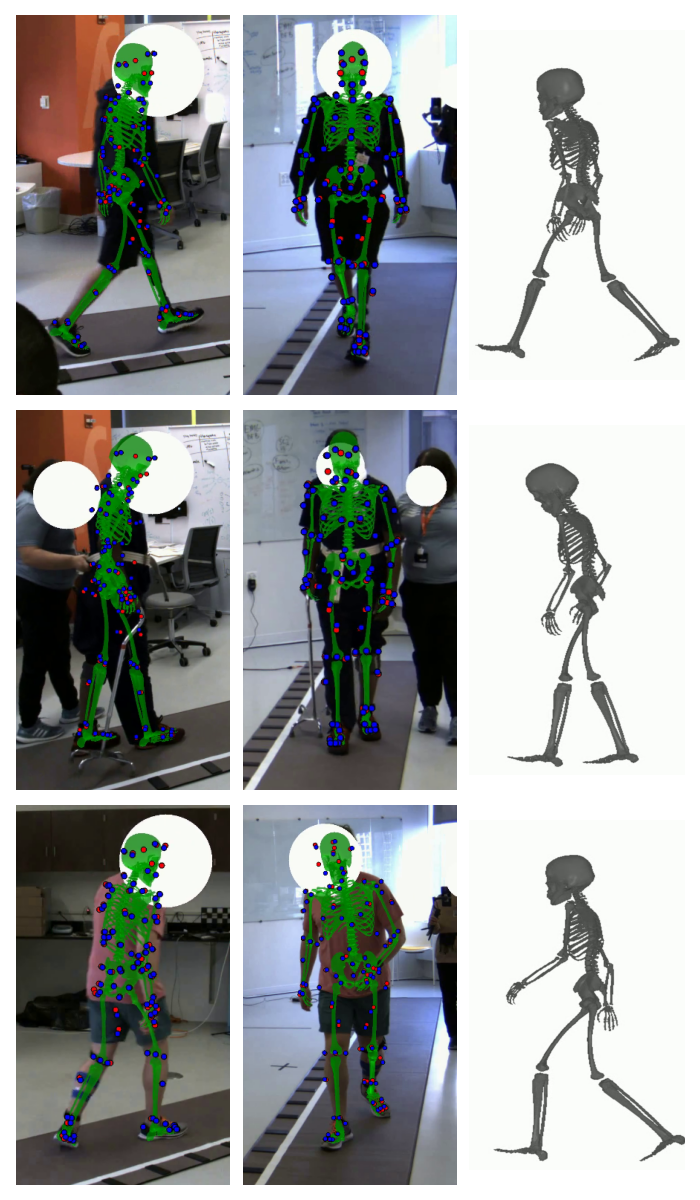}
    \caption{Example biomechanical reconstructions of rehabilitation subjects. Each row shows a single time point from an individual, with the mesh overlay shown on the first two views. Red points correspond to the HPE keypoint locations and blue points to the model markers reprojected onto the image plane. We recommend zooming in on this figure.}
    \label{fig:overlays}
    \vspace{-2.25em}
\end{figure}

In order for markerless motion capture to produce accurate joint kinematics, the 3D keypoint trajectories must contain a sufficiently dense set of observations on the body to constrain the IK results. Most HPE algorithms trained on publicly available datasets produce no keypoints between the hip and shoulders, which limits the accuracy of tracking the hip and spine \cite{needham_accuracy_2021}. One markerless motion capture system, OpenCap \cite{uhlrich_opencap_2022}, mitigates this by using an existing biomechanics dataset and training a neural network to map from a trajectory of these sparse keypoints to the most likely set of dense keypoints. A risk of this approach is the filling in of keypoints locations not extracted from the images may bias the results to an average movement and may not generalize to rehabilitation patients who move differently than the training dataset. Another challenge of existing HPE datasets is the marker and joint definitions are not biomechanically well grounded and vary greatly \cite{rapczynski_baseline_2021}. A recent study trained an autoencoder as a ``Rosetta stone'' to translate between these different datasets, including some with a dense set of biomechanically grounded keypoints typical in marker-based tracking \cite{ghorbani_movi_2021}, and found significant benefits from this cross-dataset training \cite{sarandi_learning_2022}.

Scaling the skeleton before IK typically requires additional data to be collected while the subject stands still. This can slow down a clinical workflow, and standing upright unassisted can be challenging for many rehabilitation subjects. An approach to simultaneously scale the skeleton while performing IK was recently described \cite{werling_rapid_2022}. This algorithm uses the nimblephysics engine \cite{nimblephysics} from the same author, which extends the DART physics engine \cite{lee_dart_2018} for biomechanics.

The reconstructed 3D trajectories used for IK should also be smooth and anatomically consistent. This can be challenging when keypoints from HPE have noise or may detect a different person occluding a camera. Additionally, adding constraints like temporal smoothness or skeleton consistency can make optimization much slower to converge.  We recently compared several trajectory reconstruction approaches and found that using an implicit function, parameterized with a neural network, that maps from time to 3D locations enabled accurate reconstructions that are both smooth and anatomically plausible. This method produces smooth, anatomically consistent trajectories with detected step widths and stride lengths within 10mm of those measured with a GaitRite \cite{cotton_improved_2023}.

Our goal here was to determine how to best combine these components to compute joint kinematics. Specifically, our contributions are:
\begin{itemize} %[leftmargin=*]
    \item We show that IK using implicit trajectory reconstructions performs better than using robust triangulation.
    \item We find that using the dense keypoint set from \cite{ghorbani_movi_2021,sarandi_learning_2022} performs better than the more popular 25 keypoints.
    \item We show that bilevel optimization \cite{werling_rapid_2022} for markerless motion capture without a standing calibration trial.
    \item We optimize the biomechanical model and hyperparameters of the IK algorithm \cite{werling_rapid_2022}.
\end{itemize}

\section*{Methods}

\subsection*{Participants}

This study was approved by the Northwestern University Institutional Review Board. We recruited 25 participants from both the inpatient services and outpatient clinics at Shirley Ryan AbilityLab. This included people with a history of stroke (n=8), spinal tumor (n=1), traumatic brain injury (n=2), mild foot drop (n=1), knee osteoarthritis (n=2), and prosthetic users (n=11). Ages ranged from 27 to 78. Some participants used assistive devices including orthoses, rolling walkers, or a cane, and a few participants required contact guard assistance from someone nearby for safety. We intentionally recruited a diverse rehabilitation population both to ensure the system was robustly validated for all types of patients, and because this is the population we intend to apply this final system to. The data included 162 trials containing 2164 steps, with steps per participant ranging from 44-154, with over 1 million video frames.

%In many cases, participants were also wearing wearable sensors and were followed by another team member acquiring monocular video from a smartphone as part of the ongoing development of gait analysis algorithms that can be performed with only a smartphone \cite{cotton_posepipe_2022,cimorelli_portable_2022}.

\subsection*{Data Acquisition, Video Processing, and Trajectories}

Data acquisition, video processing, and trajectory reconstruction is similar to in \cite{cotton_improved_2023} and are summarized here.

\subsubsection*{Data Acquisition}
Multicamera data was collected with a custom system in a $7.4m \times 8m$ room with subjects walking the length of the diagonal ($11m$). We used 10 FLIR BlackFly S GigE cameras (and 8 in several early experiments), which were synchronized using the IEEE1558 protocol and acquired data at 30 fps, with a typical spread between timestamps of less than 100µs. We used a mixture of lenses including F1.4/6mm, F1.8/12m, F1.6/4.4-11mm with lenses and positions selected to ensure at least three cameras covered the participants along the walkway, although the room geometry limited coverage in the corners.

The acquisition software was implemented in Python using the PySpin interface. For each experiment, calibration videos were acquired with a checkerboard ($7 \times 5$ grid of 110mm squares) spanning the acquisition volume. Extrinsic and intrinsic calibration was performed using the anipose library \cite{karashchuk_anipose_2020}. The intrinsic calibration included only the first distortion parameter. Foot contact and toe-off timing and location were acquired using a GaitRite walkway spanning the room diagonal \cite{mcdonough_validity_2001, bilney_concurrent_2003}. 

\subsubsection*{Video processing}

Collected videos were analyzed as previously described \cite{cotton_improved_2023}. Briefly, the videos were processed use PosePipe  \cite{cotton_posepipe_2022}. Easymocap \cite{easymocap} was used for the initial multiview associations and was used to indicate the subject of interest in each recording session. In contrast to \cite{cotton_improved_2023}, in this work, 2D keypoints in the image plane were also detected using a recently developed approach that is trained on numerous 3D datasets and outputs all of these formats (MeTRAbs-ACAE) \cite{sarandi_learning_2022}. Specifically, we used the output keypoints from the MOVI dataset \cite{ghorbani_movi_2021}, which has 87  keypoint locations commonly used in optical motion capture. Importantly, this has numerous keypoints around the pelvis and torso that are not available in the more common datasets.

By default, the MeTRAbs architecture \cite{sarandi_metrabs_2021} does not produce confidence estimates for the locations. It also produces best guess location estimates for occluded joints. While it produces reasonable guesses, we wanted to discard any keypoints with low confidence.  As recommended by the author, we approximated confidence by measuring the standard deviation of each 3D joint location estimated from 10 different augmented versions of each video frame. This was converted to a confidence estimate using a sigmoid function with a half maximum at 200mm and a width of 50mm. Visual inspection of the results showed this produced low confidence for occluded joints and high confidence for observed joints. We also set the confidence to zero for any joints outside the image boundary.

\subsubsection*{Trajectory reconstruction}
We found that representing 3D keypoint trajectories with implicit functions enables better reconstructions from keypoints than triangulating (either with the direct linear transform or robustly) from the 2D observations or optimizing an explicit representation of each keypoint's location \cite{cotton_improved_2023}. Specifically, the implicit representation is a function that maps from time to the reconstructed 3D keypoint locations, $f_\theta : t \rightarrow \mathbf x_t \in \mathbb R^{J \times 3}$. This is implemented as a five-layer multi-layer perceptron (MLP) with increasing numbers of hidden units (128, 256, 512, 1024, 2048), each followed by a layer normalization and a $\mathrm {relu}$ non-linearity, followed by a final dense layer that mapped to the output size. We also used sinusoidal positional encoding \cite{vaswani_attention_2017} for $t$, which was scaled from 0 to $\pi$ over each trajectory. The positional encoding of time was concatenated to the output from each layer for the first four layers.

For each trajectory, we optimize the parameters $\theta$ to minimize a loss that includes the reprojection of the 3D keypoints onto the image and loss terms that capture high-frequency noise and variations in skeleton geometry:
$$\mathcal L = \mathcal L_\Pi + \lambda_1 \mathcal L_{\mathrm {smooth}} + \lambda_2 \mathcal L_{\mathrm {skeleton}}$$
. The reprojection loss is:
$$
\mathcal L_{\Pi} = \frac{1}{T \cdot J \cdot C}\sum_{T, J, c\in C} w_{c,t,j} \, g \left( || \Pi_c \mathbf x_{t,j} - y_{t,j,c} || \right)
$$
where $\Pi_c$ is the projection operator for camera $c$ which also includes the non-linear intrinsic distortion. $y_{t,j,c} \in \mathbb R^2$ indicates the detected keypoint location for a given time point, joint, and camera and $\mathbf x_{t,j,c}$ is the corresponding 3D keypoints.  We use a Huber loss for $g(\cdot)$, which is quadratic within 5 pixels and then linear with the slope further reduced after 10 pixels of error to be more robust to outliers. The weights applied to Huber loss, $w$, are computed by a robust triangulation algorithm \cite{roy_triangulation_2022, cotton_improved_2023} that captures the consistency of each camera with the keypoint triangulation from the other cameras and down weights outliers.
We also compared using trajectories computed with robust triangulation \cite{roy_triangulation_2022, cotton_improved_2023}, which does not include any  smoothing or skeletal consistency parameters.

\subsection*{Biomechanics from inverse kinematics}

Given a biomechanical model and a set of marker locations, IK estimates the joint angles \cite{winter_biomechanics_2009}. This typically requires estimating three sets of parameters: anthropomorphic parameters for the skeleton size, marker offsets for that individual, and a sequence of poses. We performed this with the nimblephysics library \cite{nimblephysics} and specifically with their implementation of a dual optimization process that solves for all the parameters at the same time from a set of dynamic trials \cite{werling_rapid_2022}. The results of this algorithm are skeleton scaling parameters, $\mathcal S$, marker offsets, $\mathcal O \in \mathbb R^{87 \times 3}$, and kinematics as joint angle trajectories for the $i^{\mathrm {th}}$ trial: $\mathbf p^{(i)}_t \in \mathbb R^{40}$.

The skeleton model provides a forward kinematic function to determine the predicted marker locations from the IK results and a calibrated model: $\mathbf x^{(i)}_t = \mathcal M(p^{(i)}_t, \mathcal O, \mathcal S)$. We measured the consistency between these predicted model-based marker locations and the HPE keypoint detection.

\subsection*{Performance Metrics}

\subsubsection*{Geometric Consistency}
We measured the geometric consistency between the model-based marker trajectories and the detected keypoints, based on the distance between the reprojected model-based markers and the detected keypoints. To quantify this, we computed the fraction of the points below a threshold number of pixels, conditioned on being greater than a specified confidence interval. 
$$\delta_{t,j,c} = || \Pi_c \mathbf x_{t,j} - y_{t,j,c} ||$$
$$q(d,\lambda) = \frac{\sum (\delta_{t,j,c} < d)(w_{t,j,c} > \lambda)}{\sum w_{t,j,c} > \lambda}$$
Below, we report $GC_d=q(d, 0.5)$ with $d=5$ pixels.

\subsubsection*{Residual marker errors}
We measured the average distance between the 3D keypoint trajectories computed from the image keypoints and the marker locations computed from the biomechanical models after the IK solution.

\subsubsection*{Joint Limit Violations}
Using the soft, regularized joint limits allows the model to exceed the limits, which sometimes included  anomalous poses with joints having more than 180$^\circ$ offsets. We computed the fraction of time the model fits violated the joint limits and when it exceeded them by 50\% and 100\% ($v_0$, $v_{50}$, $v_{100}$, respectively). 

\subsubsection*{Pose Noise}
We measured the high-frequency noise in the IK results as: $n=\frac{1}{J (T-1)}\sum_j \sum_{t=0}^{T-1} (p_{t+1} - p_t)^2$.

\subsubsection*{Step Position Error}
We compared the heel position, determined by the location of the calcaneus bone, between the GaitRite and the IK results. We report the width of the error distribution as normalized IQR, $\sigma_{IQR}=0.7413 \cdot IQR(x)$.

\subsection*{Video Overlays}

We created video overlays from different camera perspectives to confirm the consistency of the fits with the underlying movements. Video overlays were created by extracting the mesh transformation to world coordinates for each body shape and time step from nimblephysics model. These were composed into a scene using the trimesh library \cite{trimesh} and rendered with PyTorch3D \cite{ravi2020pytorch3d} using the calibrated camera projection matrix. The PyTorch3D perspective camera does not support the distortion coefficient, so we applied the camera distortion to the mesh vertices prior to rendering.

\subsection*{Biomechanical model refinement}

Our biomechanical model was based upon the widely used Rajagopal model \cite{rajagopal_full-body_2016} with no muscles. We also added a three-degree of freedom neck joint.
We manually initialized the marker locations corresponding to the MOVI locations after visually inspecting numerous outputs. To refine them, we perform several iterations of fitting the model to our entire dataset and then adjusting the marker locations by the mean offsets, $\mathcal O$, across all subjects. While doing this, we also manually refined some marker positions to ensure that markers with reliable anatomic locations did not drift, such as the foot, ankle, knee, elbow, and wrist. 

\subsection*{Hyperparameter tuning}

We manually tuned several IK regularization hyperparameters including the anthropomorphic prior, ensuring the body segments have roughly consistent scaling, regularizing how far the anatomic and additional tracking markers can be adjusted, and the penalty for exceeding joint limits. Because of the time required to perform the biomechanical fits and the need to view renderings for qualitative fits, it was not possible to perform either a comprehensive grid search on hyperparameters or run an optimization algorithm.

\section*{Results}
\begin{table*}[th]

    \caption{\scriptsize Comparing the reconstruction metrics between the two keypoint sets and reconstructed method}
    \label{tab:fit_performance}
    
    \centering
\scriptsize
\begin{tabular}{llrrrrrrrr}
\toprule
 &  & Marker Err & Pose Noise & $GC_5$ & $v_{50}$ & $v_{100}$ & Step Length & Stride Length & Step Width \\
Keypoints & Method & (mm) &  &  & (\%) & (\%) &  $\sigma_{IQR}$ (mm) &  $\sigma_{IQR}$ (mm) &  $\sigma_{IQR}$ (mm) \\
\midrule
\multirow[c]{2}{*}{MOVI (87 kpts)} & Implicit Optimization & 18.680 & \textbf{0.047} & \textbf{0.573} & \textbf{2.048} & \textbf{0.291} & \textbf{13.202} & \textbf{8.895} & 10.924 \\
 & Robust Triangulation & \textbf{18.046} & 0.102 & 0.537 & 2.108 & 0.463 & 13.942 & 9.900 & \textbf{10.536} \\
\midrule
Halpe (25 kpts) & Implicit Optimization & 18.111 & 0.198 & 0.444 & 9.022 & 5.937 & 16.008 & 11.291 & 21.172 \\
\bottomrule
\end{tabular}
\vspace{-2em}
\end{table*}

\subsection*{Qualitative Results}

\begin{figure}
    \centering
    \includegraphics[width=\linewidth]{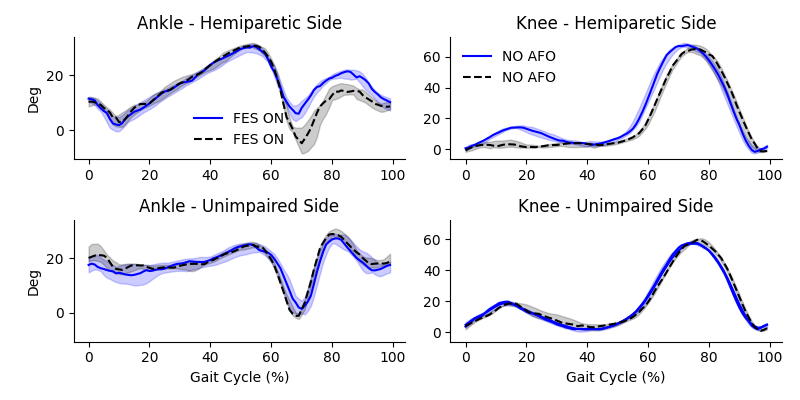}
    \caption{\scriptsize Example showing the proposed pipeline can detect clinically meaningful changes, based on multiple gait cycles being aligned and averaged. The left column shows ankle dorsiflexion during walking and the change in response to functional electrical stimulation of the peroneal nerve to improve dorsiflexion, which is seen in the trace for the impaired side. The unimpaired side shows little change, as expected. In the right column, we show changes in knee kinematics in response to an ankle-foot orthosis.}
    \label{fig:kinematic_changes}
    \vspace{-2em}
\end{figure}

The acquisition system and analysis software made it easy to perform kinematic analysis on participants seen at a rehabilitation hospital as both inpatients and outpatients, even when numerous additional people were in the room \cite{cotton_improved_2023}. Examining the overlaid results from multiple views showed good alignment between the reconstructed biomechanics and the raw image frames (Fig.~\ref{fig:overlays}). This system was also sensitive enough to pick out clinically meaningful changes in gait, as shown in Fig.~\ref{fig:kinematic_changes}. These examples used the MOVI keypoints with our optimized model and IK hyperparameters using the implicit trajectory representation. Next, we analyze these implementation details crucial for obtaining these results.

\subsection*{Dense versus Sparse Markers}
We initially compared biomechanical decomposition using a HPE algorithm using the 25 body keypoints commonly available in OpenPose and the Halpe and COCO WholeBody datasets \cite{cao_openpose_2019,fang_alphapose_2022,jin_whole_2020}. These do not include any keypoints on the torso or spine between the hips and shoulders. Similarly to previous studies, we found this produced unstable estimates of the underconstrained pelvis orientation and also produced unstable hip angles. In contrast, the dense MOVI markers from MeTRAbs-ACAE \cite{ghorbani_movi_2021,sarandi_learning_2022} produced stable results. This is quantitatively reflected in the lower levels of pose noise, lower residual marker error, and better geometric consistency with the dense markers (Table~\ref{tab:fit_performance}).

%Looked into using Augmented from OpenCap but keypoint didn't match

\subsection*{Implicit Trajectories versus Robust Triangulation}

We also compared IK fits using implicit trajectories to 3D keypoint locations that use robust triangulation. The former constrains smoothness and anatomic consistency, whereas the latter is estimated independently for each marker and time point. We found the implicit representation produced better fits with lower noise  (Table~\ref{tab:fit_performance}).
% , which account for smoothness and skeletal consistency constraints when reconstructing the keypoint trajectories, are estimated independently for each marker location and time frame

\subsection*{Model and hyperparameter tuning}

Iteratively refining the model marker positions improved the results qualitatively and by our performance metrics, such as fewer joint limit violations and greater geometric consistency between the model-based markers reprojected into the images and the detected keypoints, the residual errors between the reconstructed trajectories and the model-based trajectories.
We also found enabling soft joint limits was important, as hard limits resulted in many very implausible fits (Table~\ref{tab:hyperparameters}) 

\begin{table}[ht]
    \centering
    \caption{Impact of model iteration and joint regularization.}
    \label{tab:hyperparameters}

\begin{tabular}{lrrrrr}
\toprule
 & Marker Err & Pose Noise & $GC_5$ & $v_{50}$ & $v_{100}$ \\
Model &  &  &  &  &  \\
\midrule
Final & 18.680 & 0.047 & 0.573 & 2.048 & 0.291 \\
Early &  23.466 & 0.082 & 0.477 & 3.262 & 1.417  \\
\bottomrule
\end{tabular}

\vspace{1em}

\begin{tabular}{rrrrrr}
\toprule
$\mathcal L_{\mathrm {joint}}$ & Marker Err & Pose Noise & $GC_5$ & $v_{50}$ & $v_{100}$ \\
\midrule
5.0 & 18.680 & 0.047 & 0.573 & 2.048 & 0.291 \\
Hard & 56.251 & 0.074 & 0.295 & 0.000 & 0.000 \\
\bottomrule
\end{tabular}
\vspace{-2em}
\end{table}

\subsection*{Geometric Consistency}

In our prior work, we found the greatest geometric consistency from 3D keypoint trajectories reconstructed using an implicit representation from keypoints detected with an MMPose algorithm trained on the Halpe dataset \cite{cotton_improved_2023, mmpose_contributors_openmmlab_2020, fang_alphapose_2022}, which had a $GC_5$ of 0.54. We found the trajectories computed with an implicit representation using the MOVI keypoints and MeTRAbs-ACAE \cite{sarandi_learning_2022} substantially increased the $GC_5$ to 0.75 for the implicit trajectories. After performing the IK and generating predicted marker locations from the model this dropped to 0.57, which is still greater than the trajectories using MMPose-Halpe. 

\subsection*{Spatial Gait Parameter Errors}

We compared step length, stride length, and step width measures with respect to the calcaneus location from the IK results against simultaneous GaitRite recordings, which are shown in Table~\ref{tab:fit_performance}. These errors were between 8 and 13mm.

%\subsection*{Clinically Meaningful Kinematic Differences}
% Some participants walked in different conditions as part of their standard care. For example, using functional electrical stimulation of the peroneal nerve to increase ankle dorsiflexion during swing phase, or walking with and without an ankle foot orthosis. Our system was also sensitive to these differences (Fig XX).

\section*{Discussion}\label{discussion}

Consistent with prior work \cite{uhlrich_opencap_2022}, we found a dense set of markers, particularly in the torso and pelvis, was important for stable biomechanical IK. In particular, it quartered the noise in the estimated joint angles, greatly reduced the extreme joint angle violations, and improved the accuracy of estimated gait parameters. MeTRAbs-ACAE \cite{sarandi_learning_2022} enables training from multiple datasets and can accurately estimate these dense, biomechanically motivated marker locations from a single image. Our results demonstrate this algorithm generalizes to novel settings and images. We found that the geometric consistency was markedly greater for MeTRAbs-ACAE than other algorithms we have evaluated \cite{cotton_improved_2023}, likely because it is trained on more 3D data compared to algorithms trained on manual annotation on 2D images. 

In contrast, OpenCap \cite{uhlrich_opencap_2022} uses a keypoint augmenter to map from a time series of sparse marker data to dense marker trajectories prior to IK, using an algorithm trained on previously collected biomechanical data. We investigated using the keypoint augmenter from OpenCap \cite{uhlrich_opencap_2022} for a direct comparison, but it is not trained for the sparse keypoint set we used (Halpe \cite{fang_alphapose_2022}). We anticipate using MeTRAbs-ACAE to exact more information directly from the images will be more accurate than this filling-in approach, particularly for participants with movement patterns not represented in the training data of the keypoint augmenter. 

We found that using the implicit representation to reconstruct the trajectories for the IK step was better than using robust triangulation. This halved the noise in estimated joint angles while also improving the geometric consistency of the result. This was reassuring, as a motivation for this implicit representation was to produce better trajectories for IK with biomechanical models \cite{cotton_improved_2023}. Our rationale was that it is better to account for the HPE uncertainty and geometric structure when reconstructing the trajectory than attempting to post-process a noisy trajectory for IK. In the future, these two steps could be combined and IK approaches could be extended to run directly on the HPE outputs and consider their uncertainty and geometry while providing more accurate biomechanical constraints. Differentiable physics engines like nimblephysics make this possible, although propagating the derivatives from the implicit representation through this engine is quite complicated. Differentiable physics simulators that directly operate on the GPU designed for deep learning, such as Brax, may provide another path to integrate IK more closely with the perception layers.

Optimizing both the marker locations of the model to be more consistent with the MOVI markers \cite{ghorbani_movi_2021} and the IK hyperparameters used in nimblephysics \cite{nimblephysics} was also important to produce stable and accurate fits. In particular, enabling soft joint limits avoids unstable results seen with hard joint limits and anatomically implausible results without regularization. 
%After refining the biomechanical model, having fairly high regularization to prevent the marker locations moving far also seemed beneficial. 
However, reassuringly, we found the results were not very sensitive to the other hyperparameters.

The slight decrease in the geometric consistency of the model-based markers compared to the raw trajectories is unsurprising as the model-based estimates include more biomechanical constraints. However, the model-based $GC_5$ exceeded values we previously saw using algorithms trained on 2D datasets \cite{cotton_improved_2023}. This indicates that the multi-dataset used for MeTRAbs-ACAE both results in geometrically consistent keypoints and also biomechanically consistent keypoints that maintain their relative position on the skeleton. 
This result is despite processing each frame independently. We expect that processing videos and mapping image \emph{sequences} to keypoint sequences could utilize information about body morphology and further improve biomechanical consistency. We anticipate combining biomechanical trajectory modeling with self-supervised learning on large datasets obtained with markerless motion capture will be a fruitful avenue for training these models.

%It also suggests that the HPE algorithms do not identify anatomically consistent keypoints over time, although the $GC_5$ from the implicit representations from HPE trained on multiple 3D datasets shows substantial improvements over those trained only 2D datasets. However, both approaches process frames independently and we expect that mapping image sequences to keypoint sequences could utilize information about body morphology and improve the temporal consistency.

The MeTRAbs architecture \cite{sarandi_metrabs_2021} underlying Sarandi et al. \cite{sarandi_learning_2022} produces 2D and 3D joint locations from single images. In this work, we do not leverage the 3D locations other than to estimate the 2D keypoint confidence. This is in contrast to approaches that ``lift'' from 2D keypoints to 3D keypoints \cite{zheng_deep_2020}. Lifting approaches discard a great deal of information available from the image that is retained by \cite{sarandi_learning_2022}. An important future direction will be to compare the accuracy and stability of biomechanical fits to monocular videos from similar algorithms, or to find the minimal set of cameras required for the desired accuracy. We defer this to future work.

While \cite{sarandi_learning_2022} provides a dense set of markers for biomechanics, it is fairly sparse in the hands. This sometimes resulted in unstable estimates of the wrist and did not allow estimating finger movements. We have seen using the Halpe keypoints that we can get accurate 3D reconstruction of the hands \cite{cotton_improved_2023}. The biomechanical model we are using does not include a highly articulated hand, although they exist as separate models \cite{mcfarland_musculoskeletal_2022}. In the future, we will fuse the 3D keypoints from both algorithms and develop a biomechanical model that enables full-body biomechanical tracking, including the hand.

Our current approach has several limitations. One is that the analysis does not run in realtime. We have also not incorporated solving for dynamics or incorporating physics-based constraints, which we anticipate will further improve the accuracy of the results. Our hyperparameter optimization was not cross-validated, although given the limited number of tests, robustness to the settings, and the interpretable nature of the parameters, we do not anticipate this will have a substantial impact. There are also commercially available markerless motion capture systems available (e.g., \cite{kanko_concurrent_2020}). However, we are unable to directly compare our methodology as this proprietary system has not been described in detail. In contrast, our approach is based on publicly available algorithms.

The quantitative consistency of the model-based marker reconstruction with the detected keypoints and our qualitative impression reviewing many overlay videos indicates our approach produces accurate biomechanical kinematics. We previously found the accuracy of these trajectories for estimating spatial gait parameters was within 10mm \cite{cotton_improved_2023}, and we obtain similar accuracy after biomechanical reconstruction. The sensitivity of this approach is further supported by our ability to detect anticipated changes between walking conditions, such as bracing or functional electrical stimulation. We anticipate additional studies validating the accuracy and reliability, following the approach of other markerless motion capture systems \cite{kanko_inter-session_2021} looking at the inter-session repeatability of kinematic estimates on people without gait impairments. We did not attempt this for our current study as many of the participants have gait impairments associated with increases in gait variability, which would obscure the results.

In conclusion, advances in HPE trained on multiple datasets \cite{sarandi_learning_2022} provide dense markers for each image that are commonly used in marker-based motion capture and are sufficient to constrain biomechanical reconstructions.
Performing IK on trajectories reconstructed with an implicit representation designed to be smooth and anatomically consistent further improves the reconstruction results.
Our acquisition hardware uses cameras that only require Ethernet both for power, data transmission, and synchronization making them easy to set up in multiple locations.
Our current workflow for markerless motion capture in a rehabilitation hospital makes it easy and reliable to obtain biomechanical estimates of movement.
We hope more routine use of this technology will enable better phenotyping of patients, monitoring their response to differing interventions, and ultimately guide more effective, precision rehabilitation interventions.
We were able to bring in participants from both inpatient and outpatient clinics and perform gait analysis with several trials and conditions often in less than 10 minutes.

\printbibliography

%\addtolength{\textheight}{-30cm}

\end{document}